\documentclass[
]{ceurart}

\usepackage{listings}
\usepackage{booktabs}
\usepackage{tabularx}
\newcolumntype{Y}{>{\centering\arraybackslash}X}

\renewenvironment{quote}{%
  \list{}{%
    \leftmargin0.5cm   
    \rightmargin\leftmargin
  }
  \item\relax
}
{\endlist}

\begin{document}

\copyrightyear{2023}
\copyrightclause{Copyright c 2023 for this paper by its authors. Use permitted under Creative Commons License Attribution 4.0 International (CC BY 4.0).}

\conference{ISWC 2023 Posters and Demos: 22nd International Semantic Web Conference, November  6–10, 2023, Athens, Greece}

\title{Exploring Large Language Models for Ontology Alignment}


\author[1]{Yuan He}[%
orcid=0000-0002-4486-1262,
email=yuan.he@cs.ox.ac.uk,
]
\author[2,1]{Jiaoyan Chen}[%
orcid=0000-0003-4643-6750,
email=jiaoyan.chen@manchester.ac.uk
]
\author[1]{Hang Dong}[%
orcid=0000-0001-6828-6891,
email=hang.dong@cs.ox.ac.uk
]
\author[1]{Ian Horrocks}[%
orcid=0000-0002-2685-7462,
email=ian.horrocks@cs.ox.ac.uk
]

\address[1]{Department of Computer Science, University of Oxford}
\address[2]{Department of Computer Science, The University of Manchester}

\begin{abstract}
This work investigates the applicability of recent generative Large Language Models (LLMs), such as the GPT series and Flan-T5, to ontology alignment for identifying concept equivalence mappings across ontologies. To test the zero-shot\footnote{The term ``zero-shot'' in the context of LLMs usually refers to using the pre-trained LLMs without fine-tuning.} performance of Flan-T5-XXL and GPT-3.5-turbo, we leverage challenging subsets from two equivalence matching datasets of the OAEI Bio-ML track, taking into account concept labels and structural contexts. Preliminary findings suggest that LLMs have the potential to outperform existing ontology alignment systems like BERTMap, given careful framework and prompt design.\footnote{Our code and datasets will be made available at: \url{https://github.com/KRR-Oxford/LLMap-Prelim}}
\end{abstract}


\maketitle

\section{Introduction}
Ontology alignment, also known as ontology matching (OM), is to identify semantic correspondences between ontologies. It plays a crucial role in knowledge representation, knowledge engineering and the Semantic Web, particularly in facilitating semantic interoperability across heterogeneous sources. This study focuses on \textit{equivalence matching for named concepts}.

Previous research has effectively utilised pre-trained language models like BERT and T5 for OM \cite{he2022bertmap,amir2023truveta}, but recent advancements in large language models (LLMs) such as ChatGPT~\cite{ouyang2022training} and Flan-T5~\cite{chung2022scaling} necessitate further exploration. These LLMs, characterised by larger parameter sizes and task-specific fine-tuning, are typically guided by task-oriented prompts in a zero-shot setting or a small set of examples in a few-shot setting when applying to downstream tasks. 

This work explores the feasibility of employing LLMs for zero-shot OM. Given the significant computational demands of LLMs, it is crucial to conduct experiments with smaller yet representative datasets before full deployment. To this end, we extract two challenging subsets from the NCIT-DOID and the SNOMED-FMA (Body) equivalence matching datasets, both part of Bio-ML\footnote{OAEI Bio-ML Track: \url{https://www.cs.ox.ac.uk/isg/projects/ConCur/oaei/}} \cite{he2022machine} -- a track of the Ontology Alignment Evaluation Initiative (OAEI) that is compatible with machine learning-based OM systems. Notably, the extracted subsets exclude ``easy'' mappings, i.e., concept pairs that can be aligned through string matching. 

We mainly evaluate the open-source LLM, Flan-T5-XXL, the largest version of Flan-T5 containing 11B parameters~\cite{chung2022scaling}. We assess its performance factoring in the use of concept labels, score thresholding, and structural contexts. For baselines, we adopt the previous top-performing OM system BERTMap and its lighter version, BERTMapLt. Preliminary tests are also conducted on GPT-3.5-turbo; however, due to its high cost, only initial results are reported. Our findings suggest that LLM-based OM systems hold the potential to outperform existing ones, but require efforts in prompt design and exploration of optimal presentation methods for ontology contexts.

\section{Methodology}

\noindent\textbf{Task Definition}\hspace{0.2cm} The task of OM can be defined as follows. Given the source and target ontologies, denoted as $\mathcal{O}_{src}$ and $\mathcal{O}_{tgt}$, and their respective sets of named concepts $\mathcal{C}_{src}$ and $\mathcal{C}_{tgt}$, the objective is to generate a set of mappings in the form of ${(c \in \mathcal{C}_{src}, c' \in \mathcal{C}_{tgt}, s_{c \equiv c'})}$, where $c$ and $c'$ are concepts from $\mathcal{C}_{src}$ and $\mathcal{C}_{tgt}$, respectively, and $s_{c \equiv c'} \in [0, 1]$ is a score that reflects the likelihood of the equivalence $c \equiv c'$. From this definition, we can see that a paramount component of an OM system is its mapping scoring function $s: \mathcal{C}_{src} \times \mathcal{C}_{tgt} \rightarrow [0,1]$. In the following, we formulate a sub-task for LLMs regarding this objective.

\vspace{.2cm}

\noindent\textbf{Concept Identification}\hspace{0.2cm} This is essentially a binary classification task that determines if two concepts, given their names (multiple labels per concept possible) and/or additional structural contexts, are identical or not. As LLMs typically work in a chat-like manner, we need to provide a task prompt that incorporates the available information of two input concepts, and gather classification results from the responses of LLMs. To avoid excessive prompt engineering, we present the task description (as in previous sentences) and the available input information (such as concept labels and structural contexts)  to ChatGPT based on GPT-4\footnote{ChatGPT (GPT-4 version): \url{https://chat.openai.com/?model=gpt-4}}, and ask it to generate a task prompt for an LLM like itself. The resulting template is as follows:
\begin{quote}
\scriptsize
    Given the lists of names \textit{and hierarchical relationships} associated with two concepts, your task is to determine whether these concepts are identical or not. Consider the following:

    Source Concept Names: <list of concept names>
    
    \textit{Parent Concepts of the Source Concept: <list of concept names>}
    
    \textit{Child Concepts of the Source Concept: <list of concept names>}

    ... (same for the target concept)

    Analyze the names \textit{and the hierarchical information} provided for each concept and provide a conclusion on whether these two concepts are identical or different (``Yes'' or ``No'') based on their associated names \textit{and hierarchical relationships}.
\end{quote}
\noindent where the \textit{italic} part is generated in the second round when we inform ChatGPT parent/child contexts can be considered. Since the prompt indicates a yes/no question, we anticipate the generation of ``Yes'' or ``No'' tokens in the LLM responses. For simplicity, we use the generation probability of the ``Yes'' token as the classification score. Note that this score is proportional to the final mapping score but is not normalised. 
For ranking-based evaluation, 
given a source concept, we also consider candidate target concepts with the ``No'' answer as well as their ``No'' scores, placing them after the candidate target concepts with the ``Yes'' answer in an ascending order -- a larger ``No'' score implies a lower rank.

\section{Evaluation}

\noindent\textbf{Dataset Construction}\hspace{.2cm}
Evaluating LLMs with the current OM datasets of normal or large scales can be time and resource intensive. To yield insightful results prior to full implementation, we leverage two challenging subsets extracted from the NCIT-DOID and the SNOMED-FMA (Body) equivalence matching datasets of the OAEI Bio-ML track. We opt for Bio-ML as its ground truth mappings are curated by humans and derived from dependable sources, Mondo and UMLS. 
We choose NCIT-DOID and SNOMED-FMA (Body) from five available options because their ontologies are richer in hierarchical contexts. 
For each original dataset, we first randomly select $50$ \textit{matched} concept pairs from ground truth mappings, but excluding pairs that can be aligned with direct string matching (i.e., having at least one shared label) to restrict the efficacy of conventional lexical matching.
Next, with a fixed source ontology concept, we further select 99 unmatched target ontology concepts, thus forming a total of 100 candidate mappings (inclusive of the ground truth mapping). This selection is guided by the sub-word inverted index-based idf scores as in \citet{he2022bertmap}, which are capable of producing target ontology concepts lexically akin to the fixed source concept. We finally randomly choose 50 source concepts that \textit{do not have a matched target concept} according to the ground truth mappings, and create 100 candidate mappings for each. Therefore, each subset 
comprises 50 source ontology concepts with a match and 50 without. Each concept is associated with 100 candidate mappings, culminating in a total extraction of 10,000, i.e., (50+50)*100, concept pairs.

\vspace{.2cm}

\noindent\textbf{Evaluation Metrics}\hspace{.2cm} 
From all the 10,000 concept pairs in a given subset, the OM system is expected to predict the true mappings, which can be compared against the 50 available ground truth mappings using 
Precision, Recall, and F-score defined as:
$$
P = \frac{|\mathcal{M}_{pred} \cap \mathcal{M}_{ref}|}{|\mathcal{M}_{pred}|}, \ \ R = \frac{|\mathcal{M}_{pred} \cap \mathcal{M}_{ref}|}{|\mathcal{M}_{ref}|}, \ \ F_1 = \frac{2 P R}{P + R}
$$
\noindent where $\mathcal{M}_{pred}$ refers to the set of concept pairs (among the 10,000 pairs) that are predicted as true mappings by the system, and $\mathcal{M}_{ref}$ refers to the $50$ ground truth (reference) mappings. 

Given that each source concept is associated with 100 candidate mappings, we can calculate ranking-based metrics based on their scores. Specifically, we calculate Hits@1 for the 50 matched source concepts, counting a hit when the top-scored candidate mapping is a ground truth mapping. The MRR score is also computed for these matched source concepts, summing the inverses of the ground truth mappings' relative ranks among candidate mappings. These two scores are formulated as:
$$
Hits@K = \sum_{(c, c') \in \mathcal{M}_{ref}} \mathbb{I}_{rank_{c'} \leq K} / |\mathcal{M}_{ref}|, \ \ MRR = \sum_{(c, c') \in \mathcal{M}_{ref}} rank_{c'}^{-1} / |\mathcal{M}_{ref}|
$$

For the 50 unmatched source concepts, we compute the Rejection Rate (RR), considering a successful rejection when \textbf{all} the candidate mappings are predicted as false mappings by the system. The unmatched source concepts are assigned a ``null'' match, denoted as $c_{null}$. This results in a set of ``unreferenced'' mappings, represented as $\mathcal{M}_{unref}$. We can then define RR as:
$$
RR = \sum_{(c, c_{null}) \in \mathcal{M}_{unref}} \prod_{d \in \mathcal{T}_c} (1 - \mathbb{I}_{c \equiv d})  / |\mathcal{M}_{unref}|
$$
\noindent where $\mathcal{T}_c$ is the set of target candidate classes for a source concept $c$, and $\mathbb{I}_{c \equiv d}$ is a binary indicator that outputs 1 if the system predicts a match between $c$ and $d$, and 0 otherwise. It is worth noting that the product term becomes 1 only when all target candidate concepts are predicted as false matches, i.e., $\forall d \in \mathcal{T}_c.\mathbb{I}_{c \equiv d}=0$.

\begin{table}
    \centering
    \footnotesize
    \renewcommand{\arraystretch}{0.9} 
    \begin{tabularx}{0.88\textwidth}{X l Y Y Y Y Y Y}
    \toprule
    \textbf{System} & & \textbf{Precision} & \textbf{Recall} & \textbf{F-score} & \textbf{Hits@1} & \textbf{MRR} & \textbf{RR}\\ \midrule
    \multicolumn{2}{l}{Flan-T5-XXL} & 0.643 & 0.720 & 0.679 & 0.860 & 0.927 & 0.860 \\
    \multicolumn{2}{l}{\hspace{1mm} + threshold} & \textbf{0.861} & 0.620 & \textbf{0.721} & 0.860 & 0.927 & \textbf{0.940} \\
    \multicolumn{2}{l}{\hspace{1mm} + parent/child}
    & 0.597 & \textbf{0.740} & 0.661 & 0.880 & 0.926 & 0.760 \\
    \multicolumn{2}{l}{\hspace{1mm} + threshold \& parent/child}
    & 0.750 & 0.480 & 0.585 & 0.880 & 0.926 & 0.920 \\ \midrule
    \multicolumn{2}{l}{GPT-3.5-turbo} & 0.217 & 0.560 & 0.313 & - & - & -
    \\ \midrule
    \multicolumn{2}{l}{BERTMap} & 
    0.750 & 0.540 & 0.628 & \textbf{0.900} & \textbf{0.940} & 0.920  \\ 
    \multicolumn{2}{l}{BERTMapLt} & 0.196 & 0.180 & 0.187 & 0.460 & 0.516 & 0.920  \\
    \bottomrule
    \end{tabularx}
    \caption{\footnotesize Results on the challenging subset of the NCIT-DOID equivalence matching dataset of Bio-ML.}
    \label{tab:ncit2doid-results}


    \begin{tabularx}{0.88\textwidth}{X l Y Y Y Y Y Y}
    \toprule
    \textbf{System} & & \textbf{Precision} & \textbf{Recall} & \textbf{F-score} & \textbf{Hits@1} & \textbf{MRR} & \textbf{RR}\\ \midrule
    \multicolumn{2}{l}{Flan-T5-XXL}
    & 0.257 & 0.360 & 0.300 & 0.500 & 0.655 & 0.640 \\
    \multicolumn{2}{l}{\hspace{1mm} + threshold} & 0.452 & 0.280 & 0.346 & 0.500 & 0.655 & 0.820 \\
    \multicolumn{2}{l}{\hspace{1mm} + parent/child} 
    & 0.387 & 0.240 & 0.296 & \textbf{0.540} & 0.667 & 0.900 \\
    \multicolumn{2}{l}{\hspace{1mm} + threshold \& parent/child} 
    & 0.429 & 0.120 & 0.188 & \textbf{0.540} & 0.667 & 0.940 \\ \midrule
    \multicolumn{2}{l}{GPT-3.5-turbo} & 0.075 & 0.540 & 0.132 & - & - & -
    \\ \midrule
    \multicolumn{2}{l}{BERTMap} &
    0.485 & \textbf{0.640} & \textbf{0.552} & 
    \textbf{0.540} & \textbf{0.723} & 0.920 \\
    \multicolumn{2}{l}{BERTMapLt} & \textbf{0.516} & 0.320 & 0.395 & 0.340 & 0.543 & \textbf{0.960}\\
    \bottomrule
    \end{tabularx}
    \caption{\footnotesize Results on the challenging subset of the SNOMED-FMA (Body) equivalence matching dataset of Bio-ML.}
    \label{tab:snomed2fma-results}
    \vspace{-.3cm}
\end{table}

\vspace{.2cm}

\noindent\textbf{Model Settings}\hspace{.2cm} We examine Flan-T5-XXL under various settings: \textit{(i)} the vanilla setting, where a mapping is deemed true if it is associated with a ``Yes'' answer; \textit{(ii)} the threshold\footnote{The thresholds are empirically set to $0.650$, $0.999$, and $0.900$ for Flan-T5-XXL, BERTMap, and BERTMapLt in a pioneer experiment concerning small fragments.} setting, which filters out the ``Yes'' mappings with scores below a certain threshold; \textit{(iii)} the parent/child setting, where sampled parent and child concept names are included as additional contexts; and \textit{(iv)} parent/child+threshold setting, incorporating both structural contexts and thresholding. We also conduct experiments for GPT-3.5-turbo, the most capable variant in the GPT-3.5 series, using the same prompt. However, only setting \textit{(i)} is reported due to a high cost of this model. For the baseline models, we consider BERTMap and BERTMapLt \cite{he2022bertmap,he2023deeponto}, where the former uses a fine-tuned BERT model for classification and the latter uses the normalised edit similarity. Note that both BERTMap and BERTMapLt inherently adopt setting \textit{(ii)}. 

\vspace{.2cm}

\noindent\textbf{Results}\hspace{.2cm} 
As shown in Table \ref{tab:ncit2doid-results}-\ref{tab:snomed2fma-results}, we observe that the Flan-T5-XXL (+threshold) obtains the best F-score among its settings. While it outpaces BERTMap by 0.093 in F-score on the NCIT-DOID subset but lags behind BERTMap and BERTMapLt by 0.206 and 0.049, respectively, on the SNOMED-FMA (Body) subset. Regarding MRR, BERTMap leads on both subsets. Among Flan-T5-XXL settings, using a threshold enhances precision but reduces recall. Incorporating parent/child contexts does not enhance matching results -- this underscores the need for a more in-depth examination of strategies for leveraging ontology contexts. GPT-3.5-turbo\footnote{The experimental trials for text-davinci-003 and GPT-4 also showed suboptimal results.} does not perform well with the given prompt. One possible reason is the model's tendency to furnish extended explanations for its responses, making it challenging to extract straightforward yes/no answers. Besides, no ranking scores are presented for GPT-3.5-turbo because it does not support extracting generation probabilities. The suboptimal performance of BERTMapLt is as expected because we exclude concept pairs that can be string-matched from the extracted datasets while BERTMapLt relies on the edit similarity score.

\section{Conclusion and Future Work}

This study presents an exploration of LLMs for OM in a zero-shot setting. Results on two challenging subsets of OM datasets suggest that using LLMs can be a promising direction for OM but various problems need to be addressed including, but not limited to, the design of prompts and overall framework\footnote{This work focuses on the mapping scoring, but the searching (or candidate selection) part of OM is also crucial, especially considering that LLMs are highly computationally expensive.}, and the incorporation of ontology contexts. Future studies include refining prompt-based approaches, investigating efficient few-shot tuning, and exploring structure-informed LLMs. The lessons gleaned from these OM studies can also offer insights into other ontology engineering tasks such as ontology completion and embedding, and pave the way for a broader study on the integration of LLMs with structured data.

\bibliography{ref}

\end{document}